# Central Angle Optimization for 360-degree Holographic 3D Content

HAKDONG KIM,[1] MINSUNG YOON,[2,*] AND CHEONGWON KIM[3,*]

[1]*Department of Digital Contents, Sejong University, Seoul 05006, Korea*
[2]*Communication & Media Research Laboratory, Electronics and Telecommunications Research Institute, Daejeon 34129, Korea*
[3]*Department of Software, Sejong University, Seoul 05006, Korea*
\* *msyoon@etri.re.kr, wikim@sejong.ac.kr*

**Abstract:** In this study, we propose a method to find an optimal central angle in deep learning-based depth map estimation used to produce realistic holographic content. The acquisition of RGB-depth map images as detailed as possible must be performed to generate holograms of high quality, despite the high computational cost. Therefore, we introduce a novel pipeline designed to analyze various values of central angles between adjacent camera viewpoints equidistant from the origin of an object-centered environment. Then we propose the optimal central angle to generate high-quality holographic content. The proposed pipeline comprises key steps such as comparing estimated depth maps and comparing reconstructed CGHs (Computer-Generated Holograms) from RGB images and estimated depth maps. We experimentally demonstrate and discuss the relationship between the central angle and the quality of digital holographic content.

## 1. Introduction

Depth map estimation is essential in 3D image processing, especially in computer-generated holograms (CGH). Recently the CGHs based on 360-degree multi-viewed contents have been used for various real-time realistic display channels such as stage performances, interactions on near-eyed displays, and AR/VR/metaverse platforms. In general, a pair of RGB images and depth maps per viewpoint is required to synthesize the CGHs in applying the fast Fourier transform (FFT), the algorithm which is applicable to calculate real-time holograms. For 360-degree holographic 3D content, because a set of RGB-depth map pairs arrayed along a 360-degree arc of rotation is necessary, any loss at a specific viewpoint leads to the degrading of the quality of holographic content.

In generating the 360-degree holographic video content, there is a trade-off relationship between the computational cost of CGH and the minimum central angle which is the angular separation between adjacent viewpoints around an origin at a given radius. A pair of images, which comprise a color image & a depth map, is allocated for each viewpoint so that we can construct the 360-degree holographic 3D video content. As the central angle between the nearest viewpoints decreases, the number of viewpoints increases. Thus, the smaller the angular separation between adjacent camera's perspectives, the more detailed, realistic 3D movie can be reconstructed from 360-degree CGHs. In contrast, in terms of deep learning operations to estimate depth map information for multi-viewed holographic video content, as the unit of angular separation becomes smaller, the number of total camera viewpoints becomes larger, and then the computational cost of depth map estimation from deep learning increases, resulting in the increment of the computational quantity of CGHs. Therefore, it is necessary to search for the optimum condition concerning the central angle between adjacent viewpoints through the effective deep learning model to consider both the computational cost of CGH and the image quality of the reconstructed holographic 3D scene.

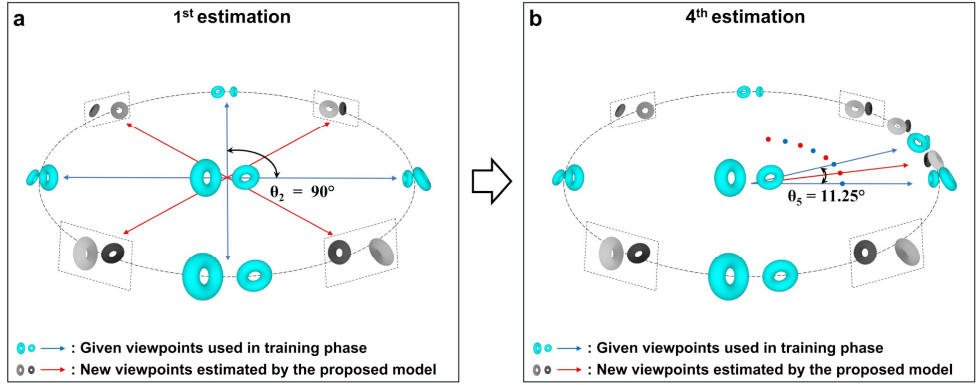

Fig. 1. Schematic diagram of 360-degree multi-viewed scenes in a rotating camera's perspective used for the proposed model. After the proposed model is trained using $2^n$ given viewpoints (blue lines or blue dots), which correspond to central angles ($\theta_n = \frac{360°}{2^n}$), the proposed model estimates $2^n$ new viewpoints (red lines or red dots) ($2 \leq n \leq 9$, where $n$ is a natural number). **a** The first estimation case: After the proposed model is trained using 4 given viewpoints, the proposed model estimates 4 new viewpoints (illustration for the case of $n = 2$, or central angle ($\theta_2$) = 90°). **b** The fourth estimation case: After the proposed model is trained using 32 given viewpoints, the proposed model estimates 32 new viewpoints (illustration for the case of $n = 5$, or central angle ($\theta_5$) = 11.25°).

In this study, we propose a novel deep learning-based model to find out the optimal conditions for the given dataset made of a pair of RGB images and depth map images per viewpoint, the conditions that are required to generate the natural 360-degree holograms and to enable users to observe the realistic holographic 3D movie or the high quality of the 360-degree reconstructed holographic 3D scene. A schematic diagram of the proposed method is shown in Fig 1. The experimental study for the proposed deep learning model comprises three steps. First, we create our own distinct dataset made of a pair of RGB images and depth map images per camera's viewpoint as a multi-viewed 3D content applicable to synthesize the 360-degree CGHs. And we train the proposed deep learning model in increasing from 4 to 512 viewpoints. Here the numbers 4 and 512 in viewpoints of the camera's perspective correspond to 90° and 0.7° in terms of the central angle between adjacent viewpoints around a central point of the given scene, respectively. The general relation between the central angle ($\theta_n$) is given by

$$\theta_n = \frac{360°}{2^n} \tag{1}$$

where $n$ is the natural number with the range of $2 \leq n \leq 9$ in the cases of the study. Second, through the proposed model, we estimate depth maps from RGB images of viewpoints not used to perform training, then synthesize CGHs using RGB images and the estimated depth maps, and reconstruct holographic 3D images from CGHs. Third, we analyze the quality of the depth map estimation and its reconstructed 3D image with respect to each case trained with varying angles.

This work contributes to the development of more efficient and realistic 360-degree digital holographic content creation processes both by presenting a quantitative standard and providing an optimization method for the angular separation of recorded viewing perspectives within a 360° around a holographic display.

## 2. Related work

A digital hologram or CGH can be typically obtained by a variety of computational algorithms. The FFT is one of the common methods to synthesize the CGH. The FFT algorithm requires input data which consists of amplitude information from RGB color images and phase information from depth maps. The acquisition of the exact depth map is critical for generating digital holographic content, especially such as realistic 360-degree digital holographic videos. Various deep learning-based methods of depth map estimation have been studied using monocular-image set, stereo-images set, or multi-viewed-image set.

### 2.1 Depth map estimation from monocular-image information

After a convolutional neural network (CNN)-based model made up of two subnet-works was proposed by Eigen [1], other approaches for monocular-image based depth map estimation were proposed, including conditional random fields [2–4], generative adversarial networks (GANs) [5,6], and a U-net model [7,8].

Also, monocular images were used as input to perform depth map estimation, aiming to create object tracking videos in autonomous vehicle's driving situations; Zhou et al [9]. proposed a method to estimate depth information from monocular videos as well as to provide situation information corresponding to the estimated depth map results. Yang et al [10]. suggested an approach to advance previous depth map estimation models so that it could derive output images including objects surface information.

### 2.2 Depth map estimation from stereo-image information

Alagoz et al. [11] presented a depth map estimation method inspired by the human visual system based on binocular disparity, or parallax between the two eyes. Joung et al. [12] proposed a CNN model to estimate depth maps by matching cost volume using an unsupervised learning approach. Garg et al. [13] and Luo et al. [14] suggested a method to operate a pixel-shifted warp using a single image input and to create the left-eye and right-eye images which then are used to estimate a depth map. Wu et al. [15] presented a GAN model with an attention mechanism for depth map estimation instead of using a disparity refinement on stereo-images.

### 2.3 Depth map estimation from multi-viewed image information

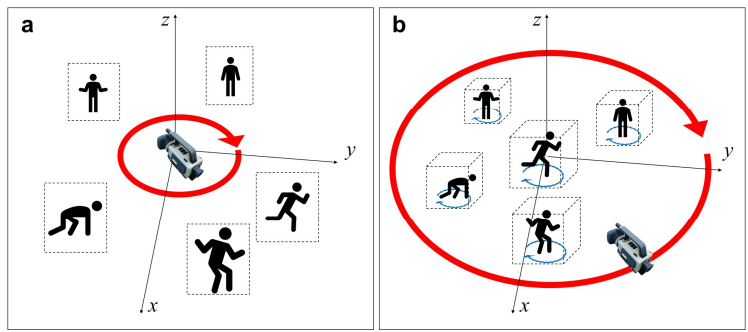

Fig. 2. Difference between **a** camera-centered environment and **b** object-centered environment. In the camera-centered environment, RGB-depth map pairs cannot be acquired for all 360-degree objects. In contrast, in the object-centered environment, RGB-Depth pairs can be acquired for 360-degree objects, which is more suitable for 360-degree holographic content.

Most of depth map estimation methods using multi-viewed images were inspired by plane-sweep methods [16], which are geometric algorithms designed to find intersecting line segments. Pei et al. [17] suggested an asymmetric U-net model for improved depth estimation from multi-viewed images for outdoor environments. Zioulis et al. [18] studied a self-supervised learning method based on geometrical formulas to estimate depth for a set of 360-degree spherical-viewed images. Feng et al. [19] presented an image set augmen-tation method designed to synthesize a 360-degree background image and a foreground image, and then to estimate depth. Both Zioulis et al. [18] and Feng et al. [19] proposed methods respectively for estimating the depth maps from 360-degree color images which are captured from a camera-centered environment, as shown in Fig 2-a.

These two methods differ from our study in that they are not suitable for 360-degree holographic content because, in a camera-centered environment, RGB-depth map pairs cannot be acquired for all directions of objects. In contrast, our proposed method is suitable for 360-degree holographic content because we use a data set of an object-centered environment that can acquire 360-degree RGB-Depth pairs for all directions of objects as shown in Fig 2-b.

### 2.4 Object-centered depth map estimation

To observe digital holographic content from within a full 360 degrees, RGB-depth map pairs must be acquired in a full 360-degree. Any missing RGB-depth map pairs at a specific viewpoint degrade the holographic content as shown in Fig 3.

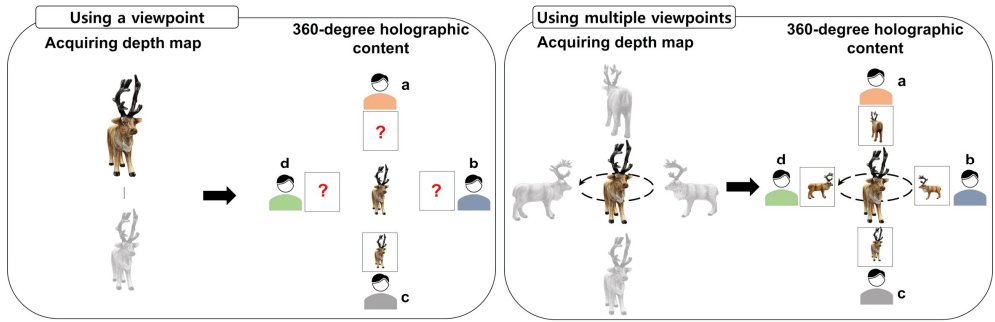

Fig. 3. 360-degree holographic content through multiple viewpoints depth map estimations. Using a viewpoint case: Only from the viewpoint of user **c** can observe holographic content because it is impossible to observe the holographic contents at the position where the depth map is missed. Using multiple viewpoints case: All viewpoints of users can observe the holographic content because the depth maps at all viewpoints have been acquired.

Recently, Kim et al. [20] proposed a method to realize a full 360-degree digital holographic content by overcoming a lack of depth map data in some viewpoints among full 360-degree viewpoints through a deep learning model to estimate the depth information for missing viewpoints. However, this method has a limitation in that depth map estimation must be performed from all viewpoints without omission because no standard has been mentioned regarding the central angle between adjacent viewpoints required for realistic CGH content. This limitation leads to the cost of depth estimation, CGH synthesis, and CGH reconstruction for more than necessary viewpoints, which must be solved to provide high-quality 360-degree holographic content. In this study, we aim to optimize and quantify the number of RGB-depth map pairs based on the central angle in an object-centered environment to represent realistic and efficient 360-degree holographic content.

## 3. Proposed method

### 3.1 Data generation

A 3D graphics software called Maya [21] is used to acquire RGB & depth map pairs for a 3D scene in an object-centered environment. To represent the accommodation effect according to depth differences in 3D space, we choose a special geometry that two objects within the scene that is mutually placed at different distances from a virtual camera's viewpoint. The central angle between adjacent viewpoints of the camera depends on the radius of the camera's rotation path. In this study, the rotational radius of the camera, that is the distance between the camera and the origin (R in Table 1), is fixed as 20 cm. The camera's setting conditions designed for image data acquisition are given in Table 1 and Fig 4.

Table 1. Supplementary Materials Supported in Optica Publishing Group Journals

| | |
|---|---|
| (A in Figure 4) Distance from virtual camera to 255 depth | 11cm |
| (B in Figure 4) Margin from depth boundary to object | 2cm |
| (C in Figure 4) Distance from virtual camera to 0 depth | 28.7cm |
| (D in Figure 4) Distance between two objects (center to center) | 8.3cm |
| (E in Figure 4) Distance from 0 depth to 255 depth | 14.2cm |
| Radius of camera rotation path (R) | 20cm |

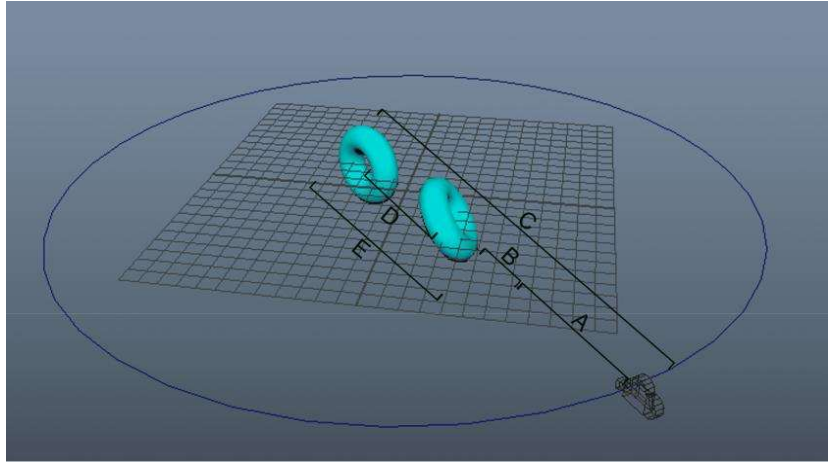

Fig. 4. Geometry of the acquisition conditions of RGB images & depth map images sets from a pair of 3D objects.

The RGB images and depth map images capturing camera is designed to rotate 360° once for about 8.5 seconds at a speed of 120 fps (frame per second). Through the camera, 1,024 RGB color images (camera-capturing of one image per 0.35° movement) are acquired during one rotation. The camera also acquires 1,024 depth maps at the same time. We divided the data set into a train set and a test set. For example, in the case of a central angle of 0.7°, 512 RGB-depth map pairs (0.7° × 512 ≈ 360°) are selected for the training process and another 512 pairs are selected for evaluation usage. Since the number of RGB-depth map pairs we acquired is 1,024, we set the minimum central angle (or the maximum number of viewpoints) in this work as 0.7° (or 512 views). Details related to the maximum central angle are mentioned in the following subsections. For 4 kinds of 3D shapes (a pair of torus, cubes, cones, and spheres), 1,024 RGB-depth map pairs from each shape are prepared for the proposed model, as shown in Fig 5.

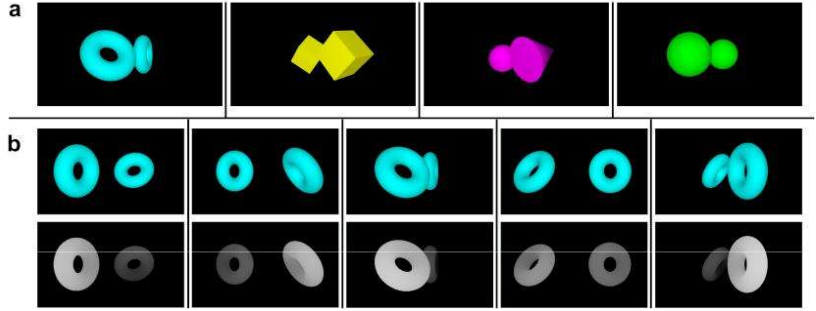

Fig. 5. Typical data sets used in the experiment: a Four kinds of 3D objects (torus, cube, cone, and sphere) are designed to construct the 360-degree, multi-viewed content. b Examples of typical data sets (torus case), each that consists of the RGB images & depth map images captured per viewpoint; each 3D scene is shot by the virtual camera at the rotational angle of 0°, 72°, 144°, 216°, and 288° on its circular track, respectively [20].

### 3.2 Model Architecture

We employed one of the existing best models for depth map estimation and simulated experiments for 360-degree, multi-viewed content, called the holographic dense depth (HDD) model, [20], which has a structure of an encoder and decoder. The HDD model is designed to estimate depth maps with high precision by using a data set, which is made up of RGB-depth map pairs taken in multi-viewed camera shots under the object-centered environment. In the HDD model, the encoder part performs feature extraction and down-sampling of input RGB color images. Each feature map is connected with a skip-connection to the decoder's up-sampling layer. Its decoder part performs up-sampling by connecting extracted features according to the size of the RGB image and estimates depth maps based on labels.

### 3.3. Process to optimize central angles

First, our experiment starts with a central angle of 90°, which is the case of $n=2$ in Figure 1. Viewpoints used for training in this case are 0°, 90°, 180°, and 270°. The proposed model was initially trained using four RGB color & depth map pairs which are captured from these four viewpoints, respectively. Then depth map estimation process is performed for a new case with four other viewpoints (for example, 45°, 135°, 225°, and 315°) using trained weights. Afterward, the proposed model was trained with a central angle of 45°, which is the case of $n=3$ in Fig 1. Eight viewpoints are used for training in this case, then the proposed model estimates depth maps for the new eight viewpoints. In the same way, we increase the number of viewpoints until $n=9$, that is until there are 512 viewpoints used for training and testing, while we trained and tested the proposed models.

## 4. Experiment Results and Discussion

### 4.1. Depth map estimation results comparison

All experiments were performed in the following hardware environment: ASUS ESC8000-G4 series with Nvidia's Titan RTX × 8. We compared depth map estimation results before CGH synthesis and reconstruction results comparison because the quality of CGH content depends on the quality of depth map estimation. Models were separately built according to the number of viewpoints corresponding to the central angle ($\theta n = 2 \leq n \leq 9$ in Fig 1.). Each model was trained using the number of viewpoints according to n and estimates depth maps for new viewpoints which are not used during training. Subsequently, we calculated the mean square error (MSE, Eq. 2) between the estimated depth maps and the ground truth. The result is shown in Fig 6.

$$MSE = \frac{1}{n}\sum_{i=1}^{n}(y_i - y'_i)^2 \qquad (2)$$

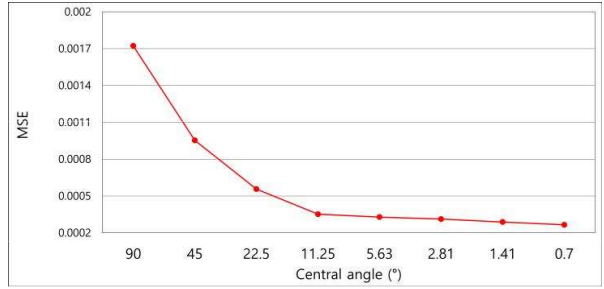

Fig. 6. Trend of the mean squared error (MSE) according to the central angle used in training the proposed model.

As shown in Figure 6, depth map estimation results showed significant improvement (about 2 times) when the size of the central angle was decreased from 90° to 45°, from 45° to 22.5°. As the size of the central angle decreased from 22.5° to 11.25°, the MSE decreased by about 1.6 times. Since then, MSE improvement has been stagnant. Therefore, we can conclude that if the central angle is less than 11.25°, there is less room for performance improvement. To additionally evaluate the quality of depth maps according to central angles, we compare them through a metric called the accuracy (ACC) [22] calculated using both ground truth depth maps and estimated depth maps. The ACC is defined as

$$Depth\ ACC = \frac{\sum_d(I \cdot I')}{\sqrt{[\sum_d I^2][\sum_d I'^2]}} \qquad (3)$$

where $I$ is the brightness of the estimated depth map, and $I'$ is the brightness of the ground truth depth map. If the estimation result and ground truth are identical, or $I=kI'$ ($k$ is a positive), $ACC=1$. If there is a mismatch between them, $0 \leq ACC \leq 1$. The results are shown in Fig 7.

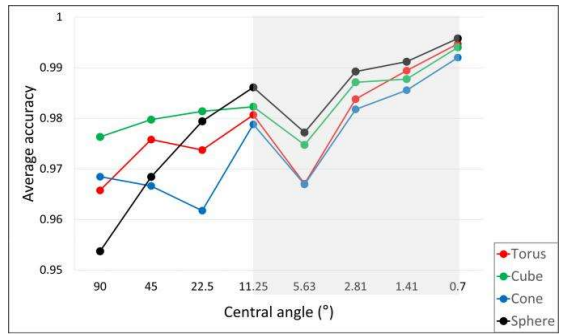

Fig. 7. Trend of the average accuracy (ACC) of depth map images according to the central angle.

For all kinds of 3D shapes we used, after ACC increased up to a condition with the central angle of 11.25°, it decreased at an angle of 5.63°, and then slowly increased again. It is true that the smaller central angle, the more improvement in the quality of the depth map. The ACC of the estimated depth map and ACC of CGH increased except for the local minimum at 5.63°, as the central angle decreased. But as the central angle passed 11.25°, the increase rate decreased relatively. To visually confirm the difference according to the central angle, the qualitative comparison result for the depth map estimation is presented in Fig 8.

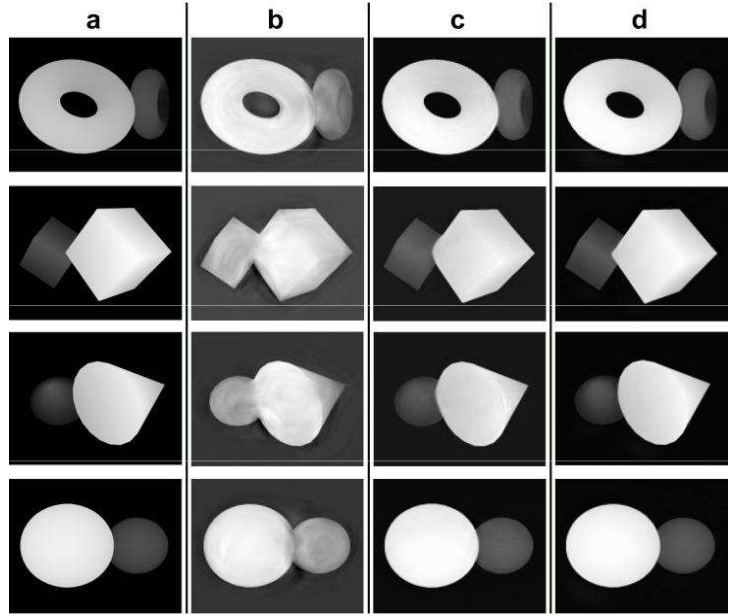

Fig. 8. Comparison of depth map estimation results according to the decrease of central angles used for training. **a** Ground truth. **b** Depth map estimation result after training in the case of using 4 given viewpoints corresponding to the central angle ($\theta$) = 90°. **c** Depth map estimation result after training in the case of using 32 given viewpoints corresponding to the central angle ($\theta$) = 11.25°. **d** Depth map estimation result after training in the case of using 512 given viewpoints corresponding to the central angle ($\theta$) = 0.7°.

The depth map estimation results of the model trained with a central angle of 90° (Fig 8-b) were inaccurate in terms of visual perception. The background and difference between distances from the camera of the two objects were inaccurately estimated. The boundary line between objects and background appears ambiguous. The boundary line where two objects overlap appears ambiguous as well. In contrast, the results of the model trained with a central angle of 11.25° (Fig 8-c) were remarkably improved, compared to the case of 90° (Fig 8-b) in terms of both estimations of the background and the overlapping area. Comparing 11.25° and 0.7°, 11.25° showed a slightly low quality. We can conclude that a smaller central angle improves the quality of the depth map. In addition, the use of 11.25° showed a significant improvement compared to 90°, whereas using 0.7° provided only a relatively small improvement compared with 11.25°.

*4.2. CGH synthesis and reconstruction results comparison*

We synthesized each CGH image from a pair of images which consist of an estimated depth map and an RGB image per view by using the FFT algorithm [23] and the Lee's encoding [24]. For the study, we generate the CGH data from the FFT algorithm [23] and the Lee's encoding scheme [24]: First, the hologram function per viewpoint is calculated by the FFT algorithm using an input set of RGB image & depth map image per viewpoint, the complex-number function which can be written as $H(x, y) = |H(x, y)|e^{i\Phi(x,y)}$ where $|H(x, y)|$ and $\Phi(x, y)$ are amplitude and phase of the hologram, respectively. Then, the Lee's encoding is used on the original hologram to get the CGH data which can be applied to directly display commercial amplitude-modulating devices such as LCD-SLMs and LCOS-SLMs [23]. Lee's representation of the hologram function can be written by

$$H(x,y) = L_1(x,y)e^{i0} + L_2(x,y)e^{i\pi/2} + L_3(x,y)e^{i\pi} + L_4(x,y)e^{i3\pi/2} \qquad (4)$$

where $L_m(x, y)$ in each component are non-negative, real-number coefficients. Each image set, used to perform the depth map estimation through the proposed model, is initially made as a resolution of 640 × 360 pixels. In consideration of the feasibility for actual holographic 3D applications such as commercial SLM (spatial light modulator) models with 4K (3840 × 2160) resolutions, we then enlarged each image's size into a resolution of 4K before generating its digital hologram fringe (CGH). To evaluate the quality of CGHs, we compare them through an ACC [22] calculated using both CGH images made from ground truth depth maps and those made from estimated depth maps. The ACC is defined as

$$CGH\ ACC = \frac{\sum_{r,g,b}(I \cdot I')}{\sqrt{[\sum_{r,g,b} I^2][\sum_{r,g,b} I'^2]}} \qquad (5)$$

where $I$ is the brightness of the CGH image from the estimated depth map, and $I'$ is the brightness of the CGH image from the ground truth depth map. If the estimation result and ground truth are identical, or $I=kI'$ ($k$ is a positive), $ACC$=1. If there is a mismatch between them, $0 \leq ACC \leq 1$. The results are shown in Fig 9.

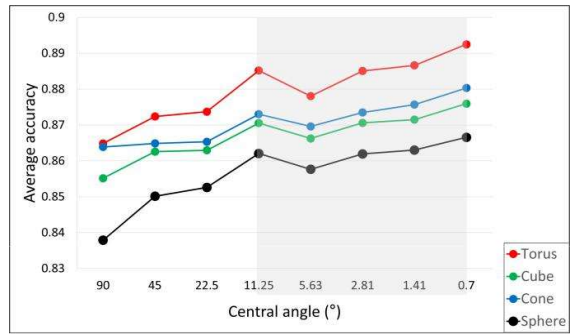

Fig. 9. Trend of the average accuracy (ACC) of CGH images according to the central angle.

We also obtained results similar to those of depth map estimation in CGH comparison experiments. For all kinds of 3D shapes we used, after *ACC* increased up to a condition with the central angle of 11.25°, it decreased at an angle of 5.63°, and then slowly increased again. This trend is similar in both depth map estimation and CGH. It is true that the smaller central angle, the more improvement in the quality of depth map estimation for 360-degree holographic content. The *ACC* of estimated depth maps and *ACC* of CGHs increased except for the local minimum at 5.63°, as the central angle decreased. But as the central angle passed 11.25°, the increase rate decreased relatively. To visually confirm the difference according to the central angle, the qualitative comparison result for the CGH reconstruction is presented in Fig 10.

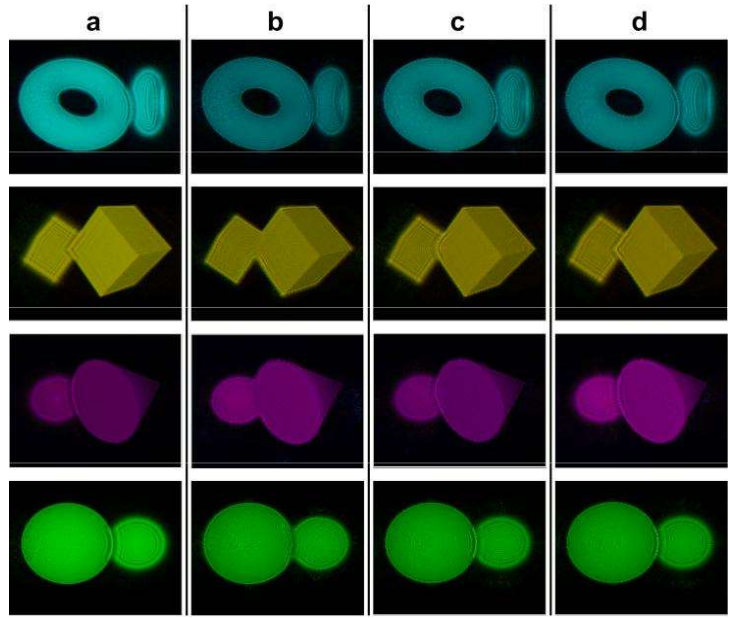

Fig. 10. Comparison of holographic 3D (H3D) images reconstructed from CGHs according to the decrease of the central angle used for training. **a** Ground truth. **b** CGH's reconstruction result after training in the case of using 4 given viewpoints corresponding to the central angle ($\theta$) = 90°. **c** CGH's reconstruction result after training in the case of using 32 given viewpoints corresponding to the central angle ($\theta$) = 11.25°. **d** CGH's reconstruction result after training in the case of using 512 given viewpoints corresponding to the central angle ($\theta$) = 0.7°.

In principle concerning holographic 3D images, observation of depth differences is one of the critical elements: When an observation camera is set to focus on one object located in the front position, the object in the front appears sharp, whereas the other object behind appears blurry. In the study, we made the experiments to optically reconstruct holographic 3D images from CGHs, and we categorize three dominant characteristic observations by qualitative comparison, as shown in Fig 10. First, the reconstruction result using the data of the model trained with a central angle of 90° is shown in (Fig 10-b); as observed in this figure, even under conditions of focusing on any object located in the front or behind, sharpness and blurring or the accommodation effect could not be distinguished. In addition, the boundary of the overlap between the two objects is not accurate. Second, it is clear that the result reconstructed using the data of the model trained with a central angle of 0.7° (Fig 10-d) is closest to the ground truth case (Fig 10-a). Third, it can be seen that the model trained with a central angle of 11.25° (Fig 10-c) realizes sufficient accommodation effect. It is also noted that the condition of 11.25° (Fig

10-c) produces the quality of the reconstructed image quite similar to that of the condition of 0.7° (Fig 10-d).

In terms of image quality based on depth map estimation, CGH's synthesis, and CGH's reconstruction results, it is found that the proposed model trained with a central angle of 0.7° achieves the best result. However, in considering both the efficiency of the data preparation process and the computational cost required to display holographic 3D content in real-time, the model trained with a central angle of 11.25° provides a more practical situation, so that the user can experience the 360-degree holographic 3D movies even with practical image quality. Also, Table 2 shows the computational time required to perform depth map learning, CGH's generation, and CGH's reconstruction, respectively, according to the central angles that were used in the training process. Here a depth map's learning time ($T$) is defined as

$$T = t \times b \times e \qquad (6)$$

where t is the time per batch, b is the number of batches, and e is the number of epochs. The number of epochs required for loss reduction depends on the central angle. Therefore, the model trained with a range of central angles between 11.25° and 0.7° used fewer epochs than that trained with a range of between 90° and 22.5°. The model trained with a central angle of 11.25° needs half the training time than the model trained with a central angle of 5.63° so the former takes about half the training time than the latter. For the same reason, the model trained with 11.25° saves about 2 hours (about 15 times) in comparison with the model trained with 0.7°. Furthermore, as shown in Table 2, the model trained with 5.63° and 0.7° did considerably increase times for CGH's synthesis and CGH's reconstruction. In terms of practical 360-degree holographic content applications, from considering both holographic 3D image quality and the content-preparation time we experimentally presented from the study, it is reasonable that the central angle is 11.25° as an optimal choice.

Table 2. Time spent on the depth map learning, the CGH generation, and the reconstruction from CGH (in the case of cone-shaped 3D objects).

| Central angle | 11.25° | 5.63° | 0.7° |
|---|---|---|---|
| Depth map learning (min'sec") | 3' 32" | 7'4" | 113'4" |
| CGH synthesis (min'sec") | 12' 10" | 97'17" | 1556'29" |
| CGH reconstruction (min'sec") | 4' 6" | 32'48" | 524'48" |

## 5. Conclusion

In the study, we proposed a deep learning-based method to search the optimal conditions for the geometric arrangement of RGB image & depth map pairs captured in an object-centered environment, so that we can get a reasonable cost efficiency to produce realistic, 360-degree holographic video content. The proposed approach consists of three steps: First, after the data set is created, the deep learning models are trained with various conditions of doubling increments from 4 to 512 viewpoints, corresponding to from 90° to 0.7° in terms of the central angle between adjacent viewpoints. Second, depth maps are estimated from RGB color images at new viewpoints not used for training, and then CGHs are synthesized through the FFT algorithm using the RGB images and estimated depth maps at the viewpoints. Holographic 3D images are reconstructed from these CGHs. Third, the quality of the estimated depth map and its reconstructed 3D image are analyzed and compared to each case trained with conditions of various central angles. We compared quantitatively/qualitatively the quality of each image from depth map estimation, CGH synthesis, and CGH reconstruction. From the experiments, we found that 11.25° among various central angles becomes the optimal condition, in considering both the displayed image quality and computational cost. In addition, we demonstrated that 360-degree holographic content generated under the very condition of 11.25° for central angle can provide image quality enough to apply for realistic 360-degree image content such as holographic 3D videos. We present a quantitative standard and optimization method based on deep learning in searching for central angles between the nearest viewpoints in the object-centered camera's environment. We expect this study to contribute to further searching the practical ways for the generation of hyper-realistic, immersive 360-degree holographic content.